\documentclass[letterpaper]{article} 
\usepackage{aaai23}  
\usepackage{times}  
\usepackage{helvet}  
\usepackage{courier}  
\usepackage[hyphens]{url}  
\usepackage{graphicx} 
\usepackage{natbib}  
\usepackage{caption} 
\usepackage{algorithm}
\usepackage{algorithmic}
\usepackage{newfloat}
\usepackage{listings}
\usepackage{array}

\usepackage{subcaption}
\usepackage{multirow}
\usepackage{booktabs}
\usepackage{bbding}
\usepackage{amsmath}
\usepackage{balance}
\newcommand{\dataset}{PM209}
\newcommand{\model}{URA}

\DeclareCaptionStyle{ruled}{labelfont=normalfont,labelsep=colon,strut=off} 
\floatstyle{ruled}
\newfloat{listing}{tb}{lst}{}
\floatname{listing}{Listing}
\title{MPMQA: Multimodal Question Answering on Product Manuals}
\author{
    Liang Zhang\textsuperscript{\rm 1},
    Anwen Hu\textsuperscript{\rm 1},
    Jing Zhang\textsuperscript{\rm 2},
    Shuo Hu\textsuperscript{\rm 2},
    Qin Jin\textsuperscript{\rm 1}\thanks{Corresponding Author.}
}
\affiliations{
    \textsuperscript{\rm 1}School of Information, Renmin University of China\\
    \textsuperscript{\rm 2}Samsung Research China - Beijing (SRC-B)\\
    \{zhangliang00,anwenhu,qjin\}@ruc.edu.cn, \{jing97.zhang,shuo.hu\}@samsung.com
}

\usepackage{bibentry}

\begin{document}

\maketitle

\begin{abstract}
Visual contents, such as illustrations and images, play a big role in product manual understanding. Existing Product Manual Question Answering (PMQA) datasets tend to ignore visual contents and only retain textual parts. 
In this work, to emphasize the importance of multimodal contents, we propose a Multimodal Product Manual Question Answering (MPMQA) task. For each question, MPMQA requires the model not only to process multimodal contents but also to provide multimodal answers. To support MPMQA, a large-scale dataset PM209 is constructed with human annotations, which contains 209 product manuals from 27 well-known consumer electronic brands. Human annotations include 6 types of semantic regions for manual contents and 22,021 pairs of question and answer. Especially, each answer consists of a textual sentence and related visual regions from manuals. Taking into account the length of product manuals and the fact that a question is always related to a small number of pages, MPMQA can be naturally split into two subtasks: retrieving most related pages and then generating multimodal answers. We further propose a unified model that can perform these two subtasks all together and achieve comparable performance with multiple task-specific models. The PM209 dataset is available at \url{https://github.com/AIM3-RUC/MPMQA}.
\end{abstract}

\section{Introduction}
\begin{figure}[t]
    \centering
    \includegraphics[width=\linewidth]{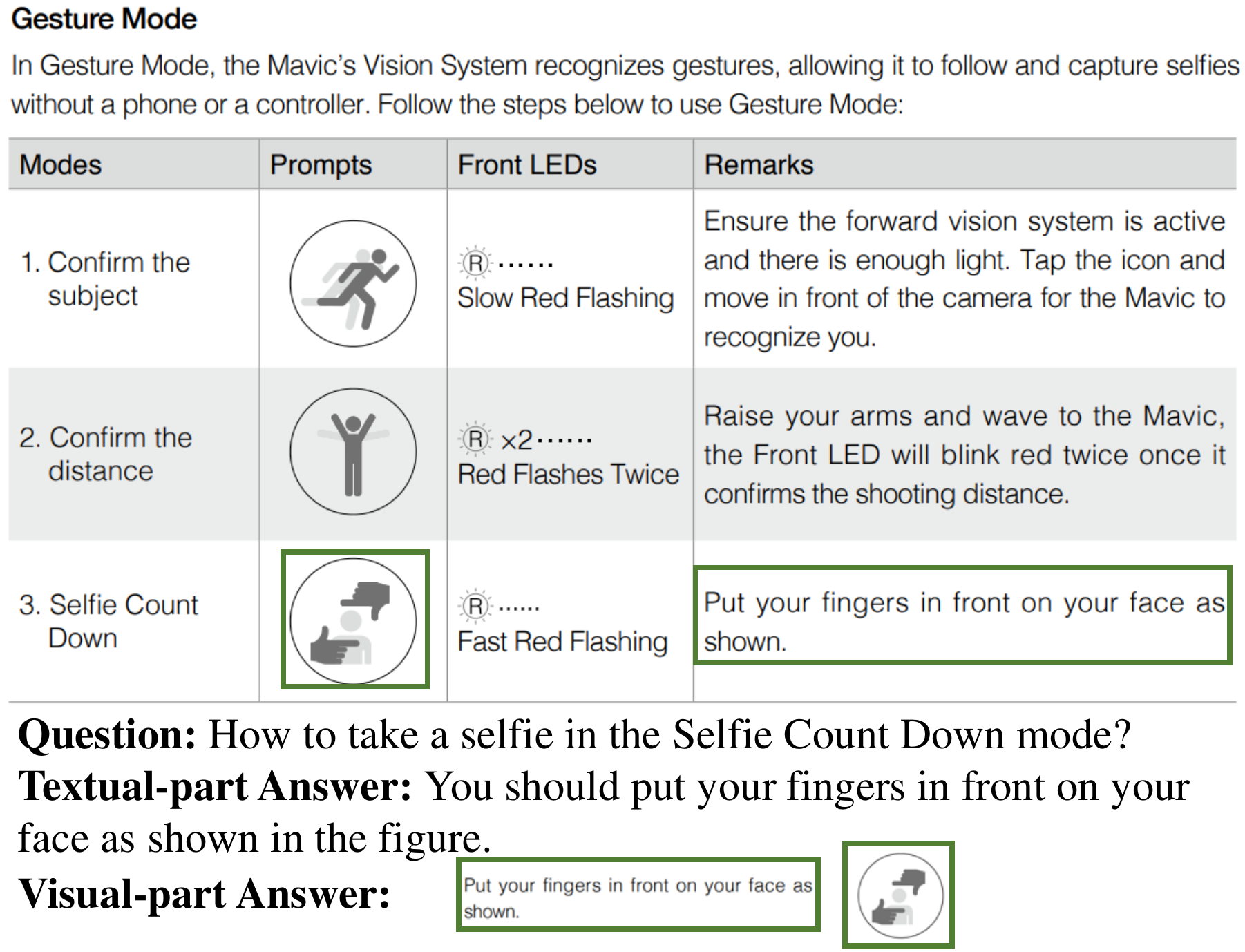}
    \caption{In this case, the gesture of the 'Selfie Count Down mode' is hard to describe using only plain text, but can be easily delivered with an image. In the MPMQA task, each question is answered with multimodal content: a textual-part answer and a visual-part answer.}
    \label{fig_1}
\end{figure}

Product manuals contain detailed descriptions of product features and operating instructions. They are often so long that it is not easy for users to efficiently find the information they are looking for. Therefore, Product Manual Question Answering (PMQA)~\cite{emanual,techqa} aims to build an AI agent on product manuals to conveniently answer user questions. PMQA leverages textual information in the manual, but ignores the visual contents, such as illustrations, tables and images, which are also important for solving user problems. As shown in Figure~\ref{fig_1}, the textual contents are insufficient to answer the question. In contrast, a multimodal answer containing both textual and visual contents can answer the question more clearly and precisely, from which users can grasp answers more effectively and efficiently. Existing Multimodal Question Answering tasks are designed to answer questions from a single web page~\cite{websrc,visualmrc} or an infographic~\cite{infographic}, which are not suitable for product manual question answering, because product manuals always contain multiple pages and most of the pages are irrelevant to the question. 
Therefore, to fill the research gap in this area,  we propose a challenging task namely \textbf{M}ultimodal \textbf{P}roduct \textbf{M}anual \textbf{Q}uestion \textbf{A}nswering (MPMQA). It requires the model to comprehend both the visual and the textual contents in an entire product manual and provide a multimodal answer for a given question.

We construct a large-scale dataset named~\dataset~with human annotations to support the research on the MPMQA task. It contains 22,021 QA annotations over 209 product manuals in 27 well-known consumer electronic brands. To support understanding of the multimodal content, we classify manual content into 6 categories (Text, Title, Product image, Table, Illustration, and Graphic). Each question is associated with a multimodal answer which is comprised of two parts: a textual part in natural language sentences, and a visual part containing regions from the manual.
Table~\ref{tab:compare_with_pmqa} shows the basic comparison between~\dataset~and existing PMQA datasets~\cite{emanual}. The scale of~\dataset~ is larger than existing PMQA datasets in terms of brands, manual numbers, and QA pairs.

Considering most pages are irrelevant to a given question, it is natural to split the MPMQA task into two subtasks: firstly retrieving the most relevant pages and then generating answers with detailed information. Thus a straightforward solution for MPMQA is to apply two task-specific models. However, both page retrieval and answer generation require the model to correlate multimodal manual contents with the question. It is possible to have both subtasks benefit from each other. Therefore, we propose the \textbf{U}nified \textbf{R}etrieval and Question \textbf{A}nswering (\textbf{\model}) model that performs these two steps with shared multimodal understanding ability. Specifically,~\model~uses a shared encoder to encode the multimodal page in the retrieval and question-answering tasks. Based on multitask learning, the~\model~model achieves comparable performance with multiple task-specific models.

Our contributions are summarized as follows:
\begin{itemize}
    \item We propose the novel MPMQA task, which requires the model to understand multimodal content in the product manual, and answer questions with multimodal outputs.
    \item We construct a large-scale dataset~\dataset~to support MPMQA. It contains not only semantic labels for manual contents, but also multimodal answers for questions. 
    \item For the MPMQA task, we design a unified model named URA that can both retrieve relevant pages and generate multimodal answers. It achieves comparable results with multiple task-specific models.
\end{itemize}
\begin{table*}[htbp]
\centering
\begin{tabular}{@{}lccccl@{}}
\toprule
Dataset & \# Manuals & \# Brands & \# QA pairs & Multimodal content & Answer type \\ \midrule
S10 QA & 1 & 1 & 904 & \XSolidBrush & Extractive \\
Smart TV/Remote QA & 1 & 1 & 950 & \XSolidBrush & Extractive \\
\dataset~(ours) & 209 & 27 & 22,021 & \Checkmark  & Multimodal \\ \bottomrule
\end{tabular}
\caption{Comparison between other question answering datasets on product manuals.}
\label{tab:compare_with_pmqa}
\end{table*}

\section{Related Works}

\subsection{Product Manuals Question Answering}
To build an automatic question-answering system, existing works explore constructing datasets based on product manuals. TechQA~\cite{techqa} collects 1400 user questions from the online forums and annotates the corresponding answers from IBM technical documents. For each question, TechQA annotates a single text span answer in the documents, similar to the strategy in SQuAD~\cite{squad,squad2}. Nandy et al. \cite{emanual} propose S10 QA and Smart TV/Remote QA datasets. They extract multiple text spans from the two Samsung device manuals to answer each question. 
These works leverage textual contents in manuals to build automatic QA systems, but ignore crucial vision information. In this work, we propose the MPMQA task, which requires models to understand both text and vision information to generate multimodal answers. Besides, our dataset~\dataset~is much bigger than the aforementioned datasets in terms of the number of products and the number of question-answering pairs. 

\subsection{Multimodal Question Answering}
Many efforts have been made to answer questions from a multimodal context. 
TextVQA~\cite{textvqa}, ST-VQA~\cite{st-vqa}, and EST-VQA~\cite{est-vqa} explore question answering on the image with scene texts. They typically require the model to extract correct scene text according to the question. ManyModalQA~\cite{manymodalqa} and MultiModalQA~\cite{multimodalqa} reason across text, tables and images from Wikipedia. DocVQA~\cite{docvqa} performs question answering on industry documents.  VisualMRC~\cite{visualmrc}, WebSRC~\cite{websrc}, WebQA~\cite{webqa} and DuReader$_{vis}$~\cite{dureadervis} require comprehension on web pages. InfographicVQA~\cite{infographic} focuses on arithmetic reasoning over infographics. Different from previous multimodal inputs, the product manual is a specific domain in terms of the question type and the content.
Since product manuals usually contain detailed operation instructions for a specific device, the questions beginning with 'How to' are very common~\cite{emanual}, while this type of contents and questions rarely occur in general domain datasets. Moreover, the answers in the above-mentioned works are all in text format, including text span, multi-choice, and generative sentences. Multimodal answers are less studied in the existing literature. MIMOQA~\cite{mimoqa} explores incorporating a Wikipedia-sourced image as a part of the answer.
Apart from the domain difference, the setting in MIMOQA is rather ideal, as it assumes all text answers associated to at least one complementary image. This assumption does not hold in product manuals. The visual-part answer in MPMQA is very diverse, not restricted to images. It can also be regions like titles and tables.
Moreover, most aforementioned works search for answers within a single document or web page. However, in the real scenario of PMQA, the target pages are not given in advance, and models have to locate relevant regions by themselves from an entire manual. To better fit the real application scenarios, our MPMQA task is designed to answer a question according to a complete manual rather than a single page, which is much more challenging than previous works.

\section{MPMQA Task and~\dataset~Dataset}
This section first presents a formal definition of the MPMQA task, and then describes the detailed process of constructing the~\dataset~dataset.
\subsection{MPMQA Task Definition}
\paragraph{\textsc{Task} (MPMQA).}
Given a question $Q$ and an n-page product manual $M=\{P_i\}_1^n$, where $P_i=\{r_{i1},\dots,r_{ik}\}$ refers to a page in $M$ and $r_{ij}$ represents a semantic region in $P_i$, the model produces a multimodal answer $A=(T,R)$ containing two parts, with $T$ as the textual-part answer in natural language sentences and $R=\{r_i\}_1^m$ as the visual-part answer consisting of multiple semantic regions. 

\noindent Since almost all questions are relevant to a very small number of pages in a manual, the MPMQA task can be naturally split into the following two subtasks: 

\paragraph{\textsc{Subtask I} (Page Retrieval).} Given a question $Q$ and an $n$-page product manual $M=\{P_i\}_1^n$, the model finds the smallest subset $\{P_{(i)}\}_1^k$ that contains the answer of $Q$.

\paragraph{\textsc{Subtask II} (Multimodal QA).} Given a question $Q$ and $k$ relevant pages $\{P_{(i)}\}_1^k$, the model generates a multimodal answer $A=(T,R)$ as defined in the \textsc{Task} MPMQA.

\subsection{\dataset~Dataset Construction}
We construct the~\dataset~dataset to support the MPMQA task. We first collect a set of product manuals. Crowd workers from the Maadaa Platform\footnote{\url{https://maadaa.ai}} then annotate the semantic regions $r$ for each page $P$ in the manuals. After that, the OCR words $W=\{w_i, b_i\}_1^n$ inside each semantic region $r$ are automatically extracted. Finally, crowd workers create (question, multimodal answer) pairs $(Q,A)$ based on the content of each manual.
All crowd workers who participated in this project are proficient English speakers.

\noindent\textbf{Product Manual Collection.} We collect 209 English product manuals in total from well-known consumer electronic brands. These manuals cover 27 brands and 90 categories. 

To ensure the diversity, we only keep the longest manual for the products in the same series. All manuals are born-in-digital PDF files and we render each page into image. We manually remove pages that are not suitable for posing questions, such as empty pages and cover pages, and ensure that all manuals in~\dataset~contain not less than 10 valid pages.

\noindent\textbf{Semantic Region Annotation.} Thirteen crowd workers are recruited to annotate the semantic regions $r_i$ of each page in the product manuals. Two crowd workers then further validate the annotations. A semantic region consists of a bounding box $b_i$ and a semantic label $c_i$. 
We define six semantic regions as follows. The example of these semantic regions can be found in the appendix.
\begin{itemize}
\setlength{\itemsep}{0pt}
\setlength{\parsep}{0pt}
\setlength{\parskip}{0pt}
    \item \textbf{Text.} The body paragraphs that convey major textual information in the product manual. 
    \item \textbf{Title.} The words summarize or indicate the section of the whole page or nearby paragraph, Titles typically consist of a few words and have different fonts than the words in the paragraph (e.g.larger size, in bold or different color).
    \item \textbf{Product image.} Product relevant images in the manual, including the picture of product, operating interface, and components of the product etc. Product irrelevant images such as decorative drawings are not included.
    \item \textbf{Illustration.} Visually rich regions to describe a particular function,  operation, and purpose of the product. They usually but not always consist of a combination of a product image and a surrounding text notes. 
    \item \textbf{Table.} Regions that convey the information of text in a row-column format. 
    \item \textbf{Graphic.} Visually rich regions indicating the name and position of a product component. It typically consists of a product image, some surrounding texts, and indicators (lines, arrows, and serial numbers) that align the names in the text regions with positions in the product image.
\end{itemize}

To reduce the burden of human annotation, we leverage PyMuPDF~\cite{pymupdf} to automatically extract bounding boxes of paragraphs and images in each page. We attach the 'Text' and 'Product image' labels to the paragraph and image bounding boxes produced by PyMuPDF respectively. The crowd workers then modify these initial bounding boxes and generate the above-mentioned semantic regions. The modification options include moving, resizing, relabeling categories, creating, and deleting. 

\noindent\textbf{Word Extraction.} Since the product manuals are born-in-digital, we automatically extract OCR words $\{w_i, b_i\}_1^n$ in each region through PyMuPDF~\cite{pymupdf}.

\noindent\textbf{QA Annotation.} Twenty crowd workers are recruited to create (question, multimodal answer) pairs $(Q,A)$ for each product manual. Considering the large cognitive load for reading the entire manual, and the fact that a question is usually only relevant to a few pages, we divide the entire product manuals into groups, and each group contains consecutive 5 pages. Crowd workers focus on each group and pose two questions for each page. For each question, they create a multimodal answer, which consists of two parts: the textual part $T$ that is written to describe the answer in natural language sentences, and the visual part $R=\{r_i\}_1^m$ that is selected from the semantic regions. To simulate the real user scenario, the annotators are encouraged to write the question in the first person, and provide textual part answer $T$ in the second person.
We do not restrict the position of the visual part answer in each group, however, crowd workers rarely pose questions across multiple pages (details in appendix).

\subsection{Statistics and Analysis}
This section presents the statistics and analysis of the proposed~\dataset~dataset.

\begin{figure}
    \centering
    \begin{subfigure}{0.5\linewidth}
    \centering
    \includegraphics[width=1\linewidth]{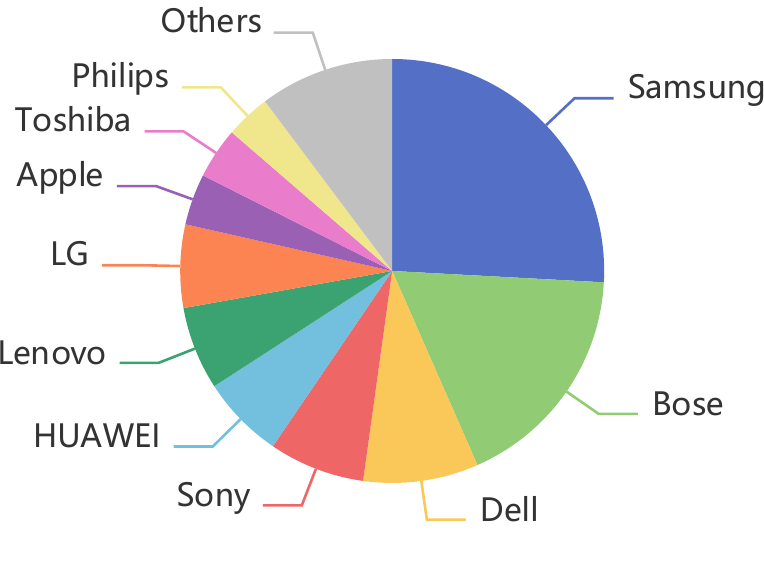}
    \label{fig:brands}
    \end{subfigure}
    \hfill
    \begin{subfigure}{0.46\linewidth}
    \centering
    \includegraphics[width=1\linewidth]{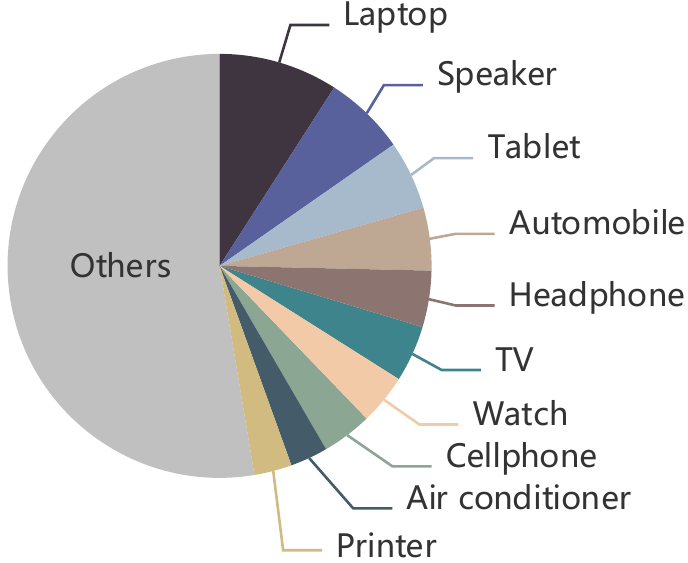}
    \label{fig:products}
    \end{subfigure}
    \caption{Top 10 brands (left) and products (right) in~\dataset.}
    \label{fig:product_brand}
\end{figure}

\noindent\textbf{Manuals.}~\dataset~consists of 209 product manuals in 27 well-known consumer electronic brands and 90 product categories. Figure~\ref{fig:product_brand} shows the top 10 products and brands. Note that the top 10 products cover less than 50\% of all manuals, which reveals that the manuals in~\dataset~are highly diverse. 
The list of all brands and products can be found in the appendix.

\begin{figure}[t]
    \centering
    \begin{subfigure}{0.48\linewidth}
    \centering
    \includegraphics[width=1\linewidth]{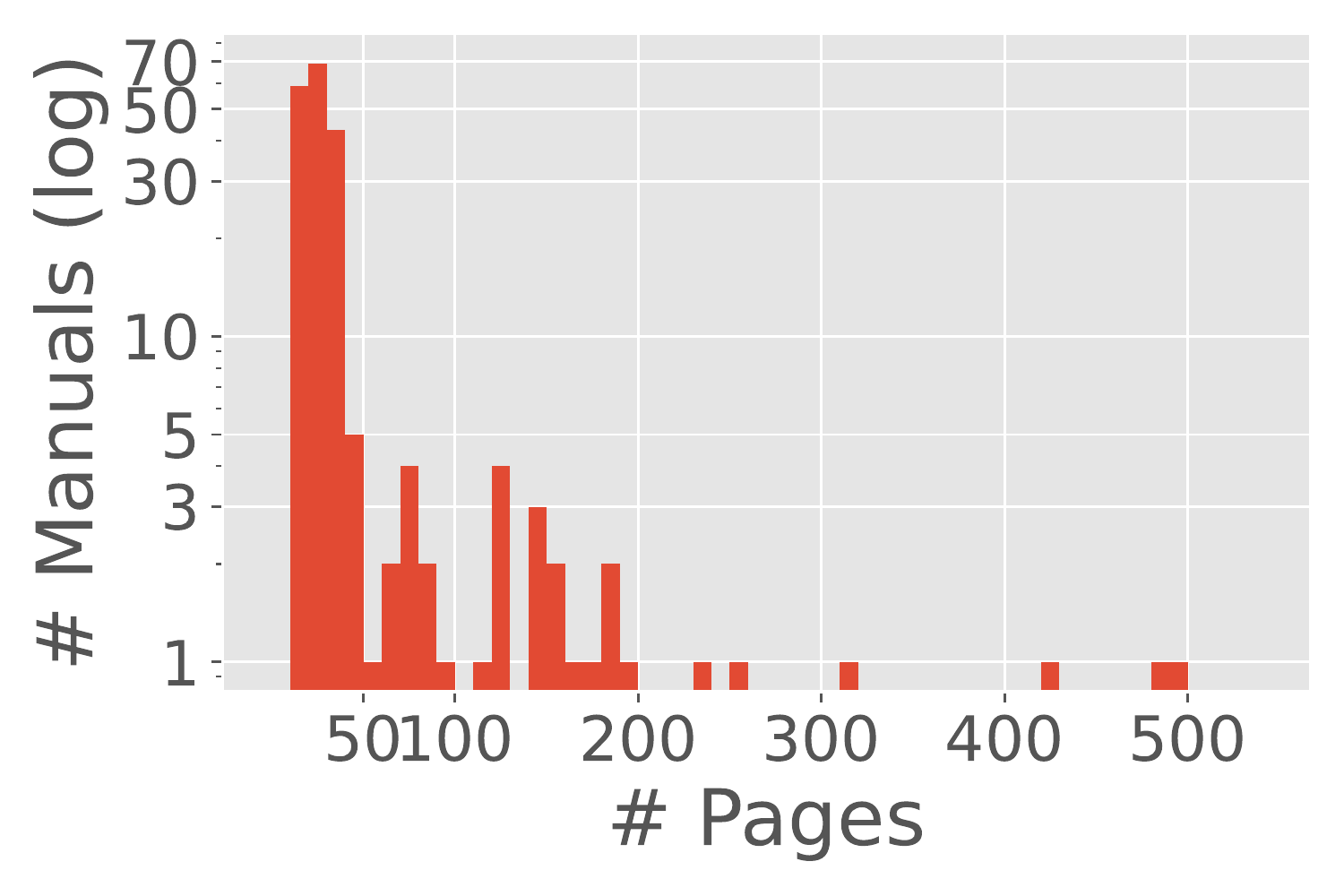}
    \caption{Count of manuals with a particular number of pages.}
    \label{fig:pages}
    \end{subfigure}
    \hfill
    \begin{subfigure}{0.48\linewidth}
    \centering
    \includegraphics[width=1\linewidth]{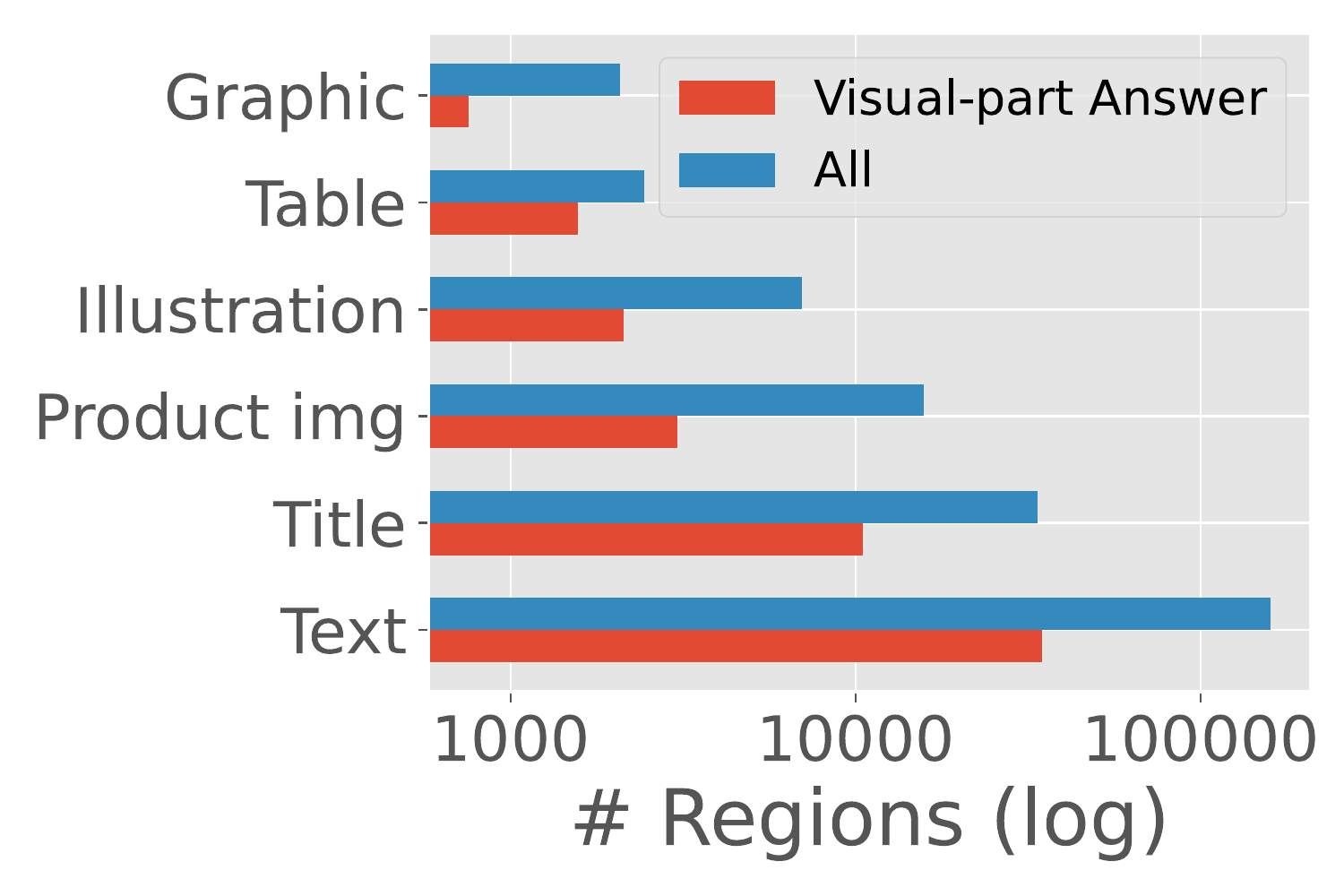}
    \caption{The number of semantic regions.}
    \label{fig:regions}
    \end{subfigure}
    \caption{Statistics over pages and regions.}
    \label{}
\end{figure}

We also analyze the distribution over the number of pages in Figure~\ref{fig:pages}. It shows that~\dataset~are also diverse in lengths, ranging from 10 pages to 500 pages. The average length of the manuals is 50.76 pages.

\noindent\textbf{Semantic regions.} Figure~\ref{fig:regions} presents the statistics of the semantic regions. We observe that product manuals indeed contain rich layout information. Specifically, 65.1\% of pages contain visually-rich regions such as product images, illustrations, tables and graphics. And 22.1\% of these regions occur in the visual-part answer.

\noindent\textbf{Questions and Textual-part Answers.} The comparison between~\dataset~and other Multimodal Question Answering datasets is shown in Table~\ref{tab:compare_with_mrc}.~\dataset~has a higher percentage of unique questions (98.46\%) and unique answers (98.35\%). It further indicates the high diversity of the~\dataset~dataset, since we avoid the appearance of similar product manuals, and both the questions and the answers in~\dataset~are specifically designed for each product. In addition,~\dataset~has the longest answer compared to other datasets, since the instruction and procedural answers can be long in product manuals.

\begin{figure}[t]
    \centering
    \begin{subfigure}{0.495\linewidth}
    \centering
    \includegraphics[width=1\linewidth]{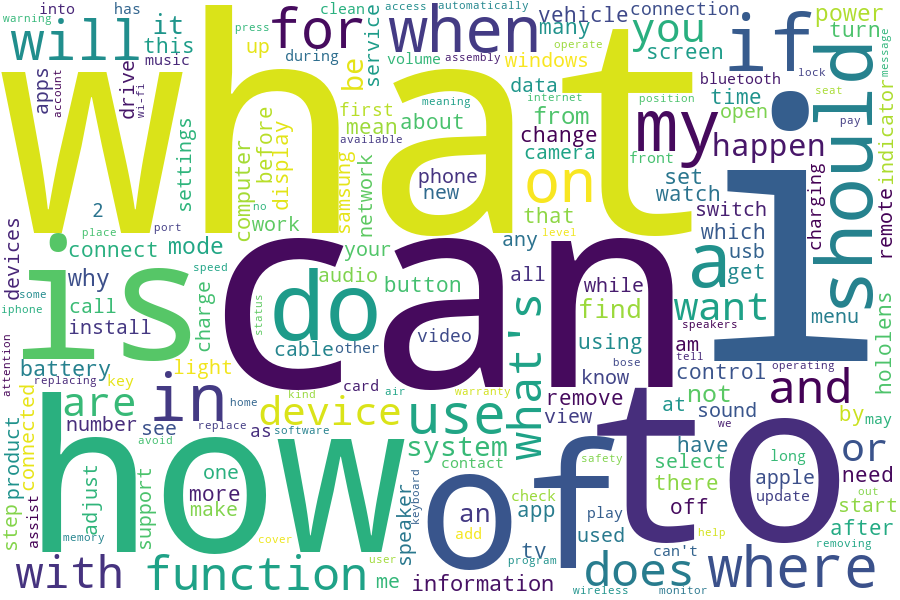}
    \label{fig:question_cloud}
    \end{subfigure}
    \hfill
    \begin{subfigure}{0.495\linewidth}
    \centering
    \includegraphics[width=1\linewidth]{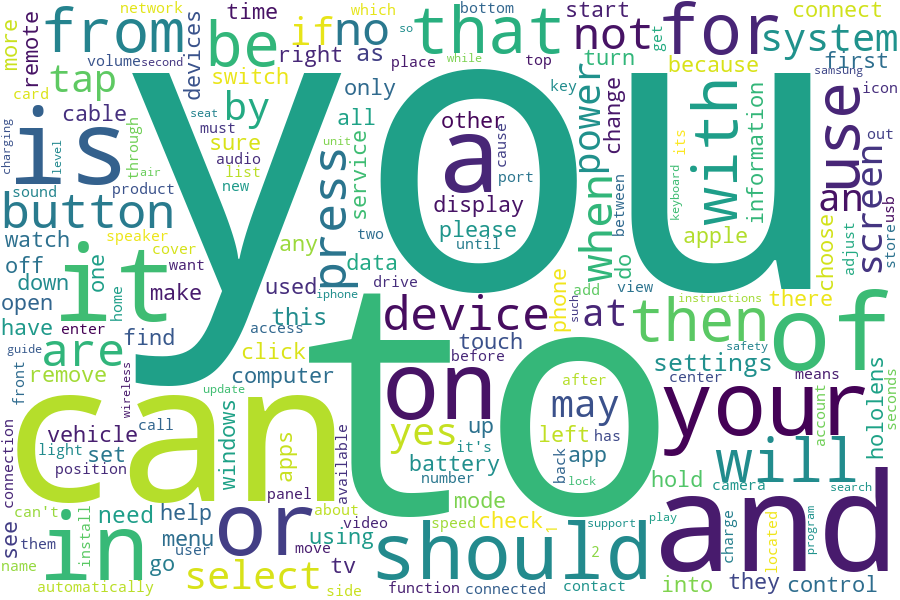}
    \label{fig:answer_cloud}
    \end{subfigure}
    \caption{Word clouds for questions (left) and textual-part answers (right) in~\dataset.}
    \label{fig:word_cloud}
\end{figure}

\begin{figure}[t]
    \centering
    \begin{subfigure}{0.495\linewidth}
    \centering
    \includegraphics[width=1\linewidth]{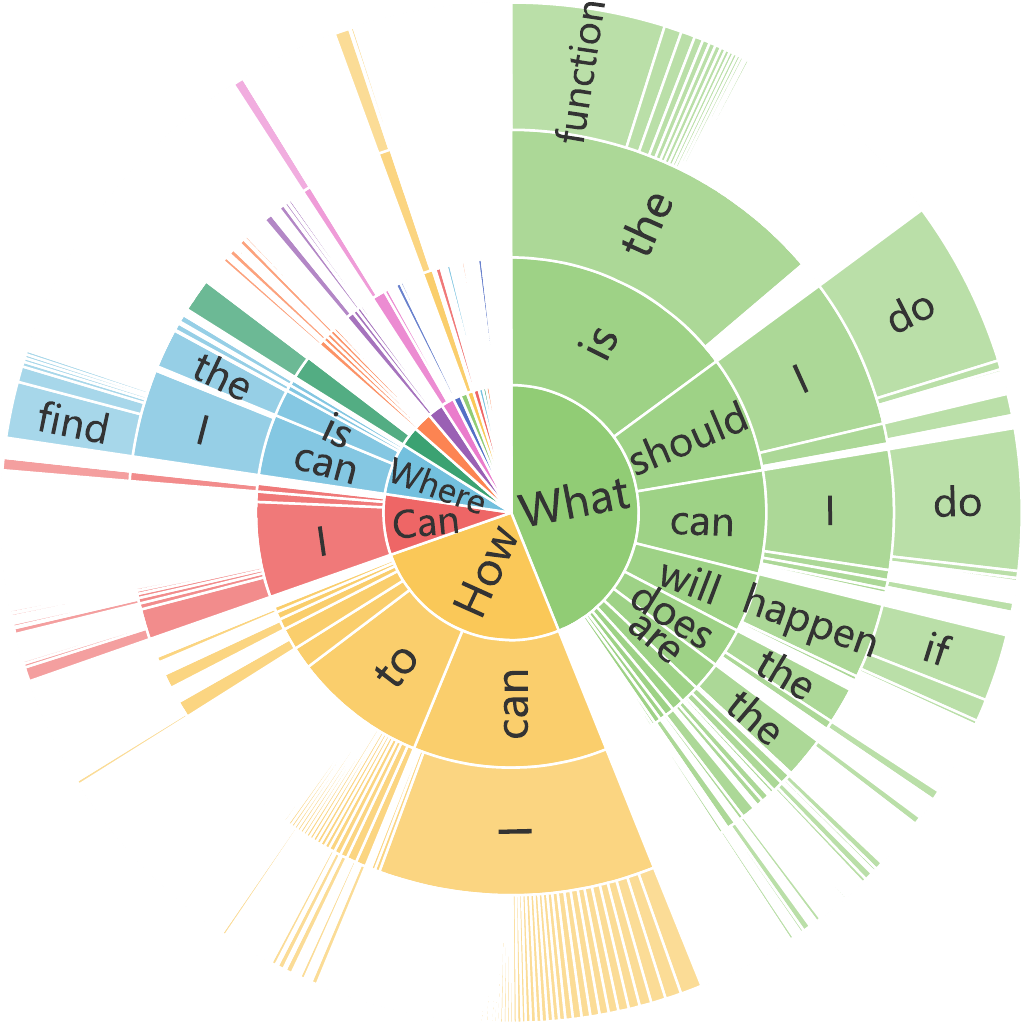}
    \label{fig:question_begin_words}
    \end{subfigure}
    \hfill
    \begin{subfigure}{0.495\linewidth}
    \centering
    \includegraphics[width=1\linewidth]{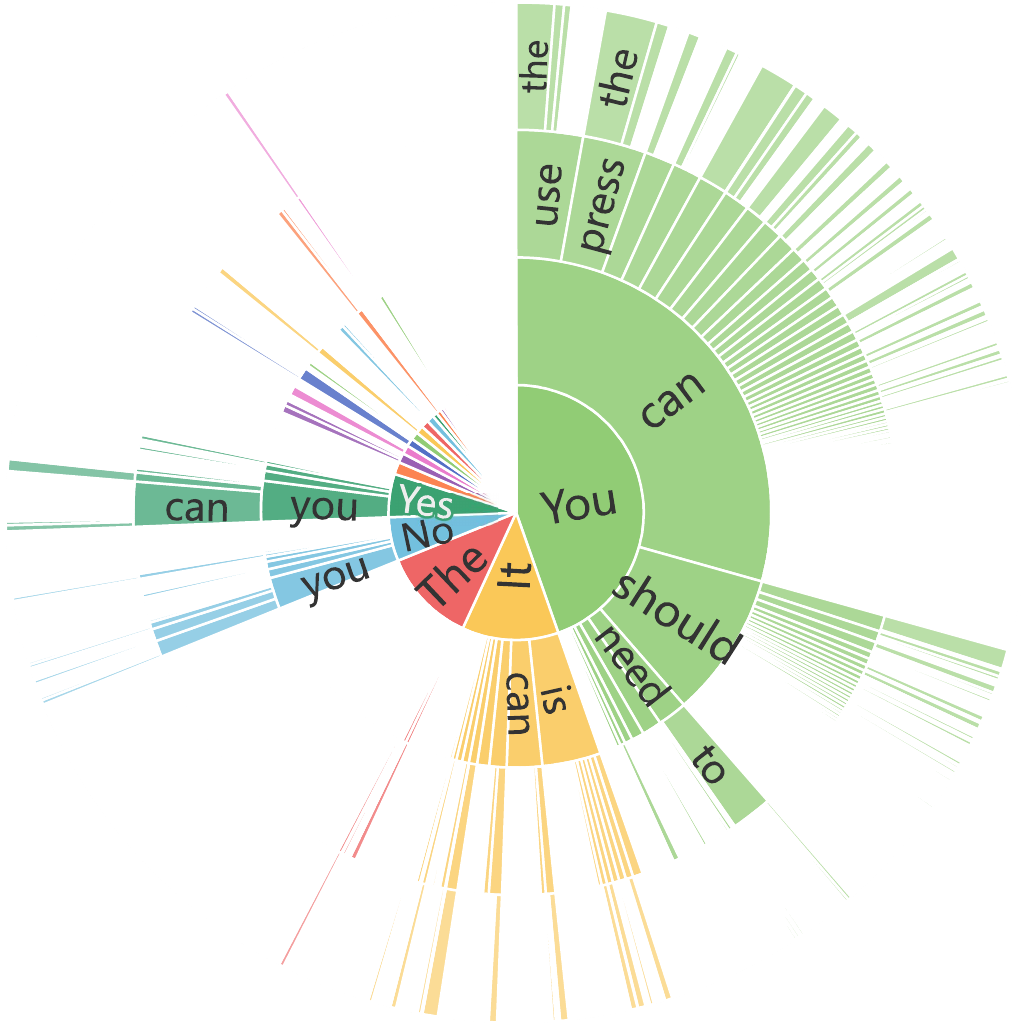}
    \label{fig:answer_begin_words}
    \end{subfigure}
    \caption{First 4-grams of questions (left) and answers (right) in~\dataset.}
    \label{fig:begin_words}
\end{figure}

Figure~\ref{fig:word_cloud} shows the word cloud of the questions and textual-part answers. We find that questions in~\dataset~contain both factual words such as 'function' and 'information', and procedural words including 'begin', 'step', and 'after'. Apart from guidance-related questions such as 'what' and 'how', the frequency of pronoun 'I' has a high frequency in the questions. Correspondingly, the word 'you' appear frequently in the answers. This is as expected since we simulate the real-world scenarios where users pose questions in the first person, while the QA system answers the questions in the second person. Figure~\ref{fig:begin_words} shows the first 4-grams of questions and answers. Most questions begin with the word 'what' (43.91\%) and 'how' (25.72\%). Questions with 'how' tend to ask about the procedural process of an operation. Questions with 'what' are typically about factual information about the product usage, except in the case of 'What should I do ...', which are also procedural questions. Besides, there are 7.71\% of questions starting with the word 'can'. These questions are usually confirming something uncertain about the product, e.g. 'Can I use this device underwater?'. Their answers usually begin with 'yes' or 'no'. 

\begin{table}[ht!]
\centering
\scalebox{0.9}{
\begin{tabular}{@{}m{60pt}m{25pt}m{28pt}m{25pt}m{28pt}m{28pt}@{}}
\hline
\multirow{2}{*}{Dataset} & \multicolumn{2}{c}{Question} & \multicolumn{2}{c}{Answers} & \multicolumn{1}{c}{Page} \\
 & \%Uniq. & Length & \%Uniq. & Length & Length \\ \hline
ST-VQA & 84.84 & 8.80 & 65.63 & 1.56 & 7.52 \\
TextVQA & 80.36 & 8.12 & 51.74 & 1.51 & 12.17 \\
DocVQA & 72.34 & 9.49 & 64.29 & 2.43 & 182.75 \\
VisualMRC & 96.26 & 10.55 & 91.82 & 9.55 & 151.46 \\
InfographicVQA & 99.11 & 11.54 & 48.84 & 1.60 & 217.89 \\
\dataset & 98.46 & 9.77 & 98.35 & 15.74 & 231.36 \\ \hline
\end{tabular}
}
\caption{Comparison of Multimodal Question Answering datasets w.r.t. uniqueness rate and the average length of questions and answers, and the average length per page.}
\label{tab:compare_with_mrc}
\end{table}

\begin{figure}[ht!]
    \centering
    \includegraphics[width=0.8\linewidth]{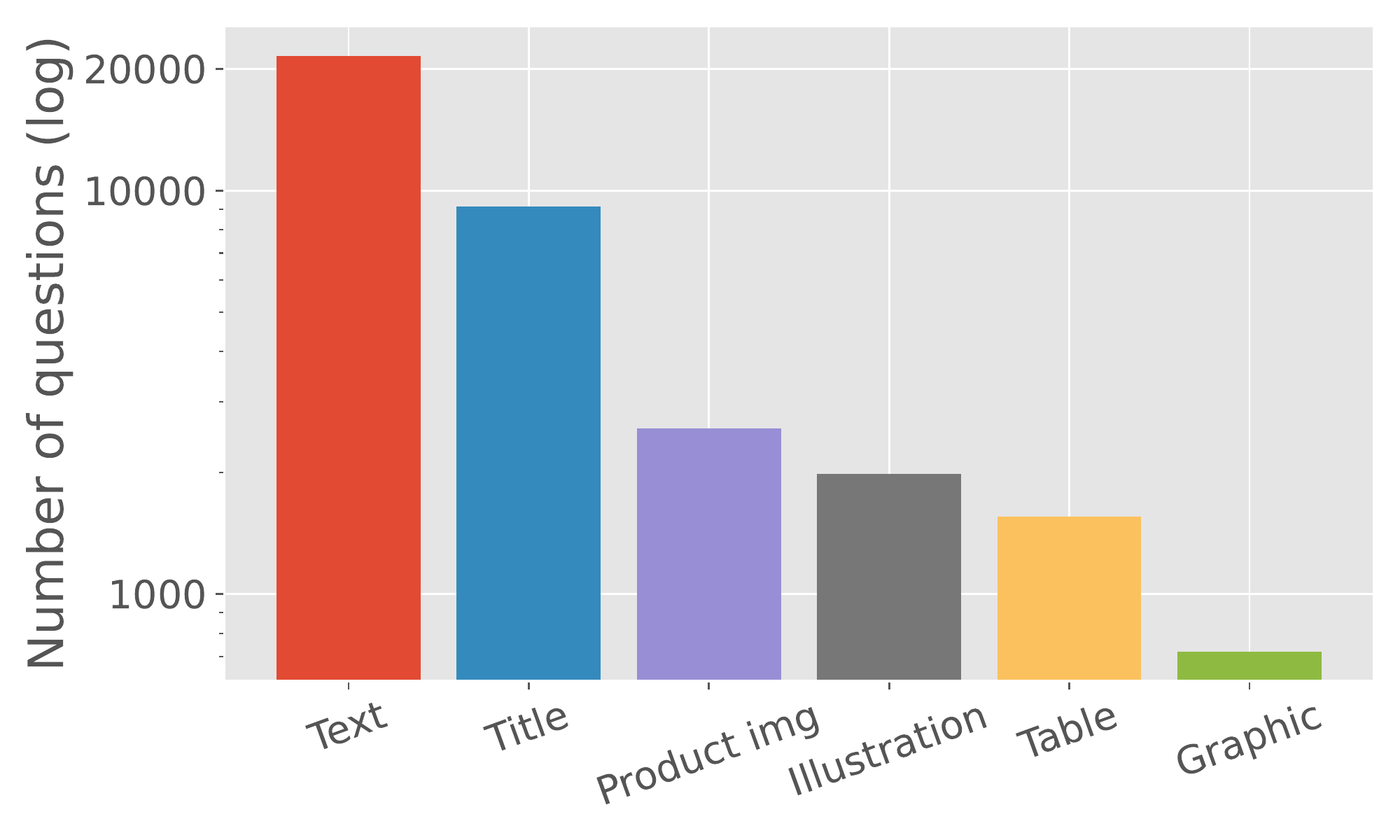}
    \caption{Visual-part answers break down by semantic labels}
    \label{fig:qa_region}
\end{figure}

\noindent\textbf{Visual-part Answers.} Apart from the textual-part answers, each question in~\dataset~is also paired with a set of regions in the product manual. These regions can be seen as complementary to understanding the text answers. Figure~\ref{fig:qa_region} shows the number of visual-part answers broken down by semantic labels. A significant portion (21.8\%) of questions include visually-rich regions (product images, illustrations, tables, and graphics) in their visual-part answers. This portion is higher than VisualMRC, in which 9.1\% of questions are relevant with visually-rich regions (picture and data). It indicates that visual components can be more important in understanding product manuals than open-domain web pages. 

\noindent\textbf{Data splits.} We divide the manuals in the~\dataset~dataset into Train/Val/Test as shown in Table~\ref{tab:splits}.
\begin{table}[t]
\centering
\begin{tabular}{lccc}
\hline
 & Train & Val & Test \\ \hline
\# Manuals & 146 & 21 & 42 \\
\# Pages & 7004 & 1011 & 2003 \\
\# QAs & 15839 & 2257 & 3925 \\ \hline
\end{tabular}
\caption{Number of samples in each data split.}
\label{tab:splits}
\end{table}

\section{Proposed Model}
\begin{figure*}[t]
    \centering
    \includegraphics[width=0.95\linewidth]{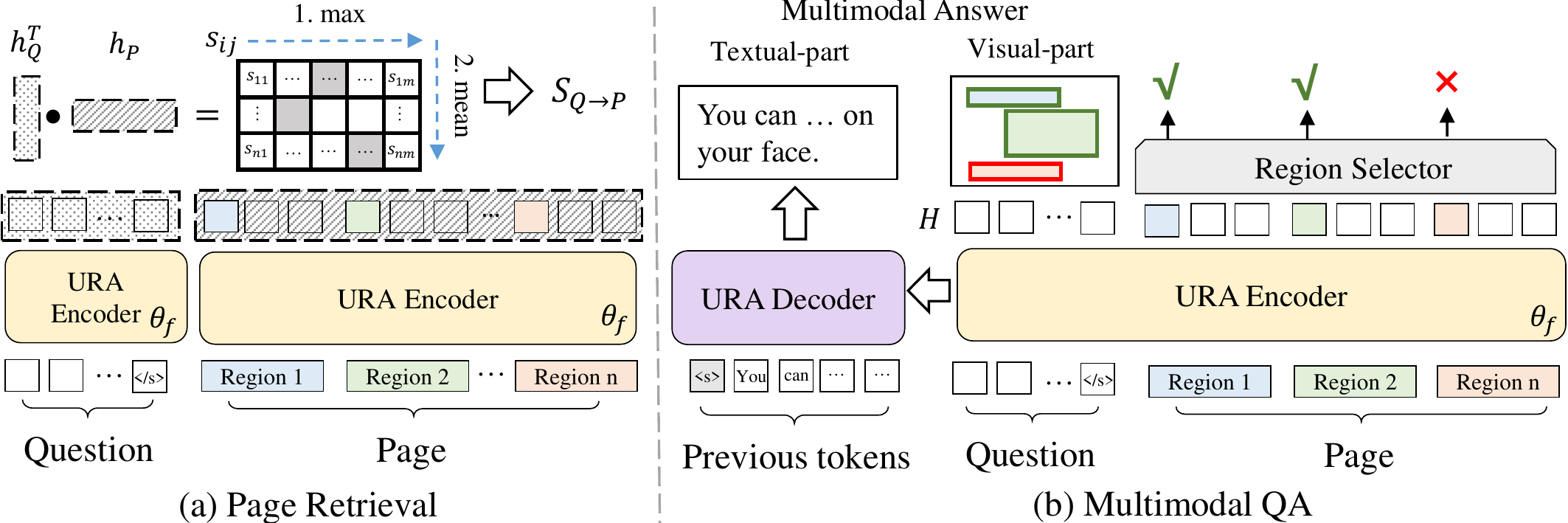}
    \caption{Overview of the Unified Retrieval and Question Answering (\model) model.}
    \label{fig:model}
\end{figure*}

We propose a \textbf{U}nified \textbf{R}etrieval and Question \textbf{A}nswering (\textbf{URA}) model for the new MPMQA task,  which can perform page retrieval and multimodal QA all together.  
As shown in Figure~\ref{fig:model}, the model consists of three key components: a~\model~Encoder, a~\model~Decoder, and a Region Selector. For the page retrieval task,~\model~encodes the questions and the pages separately, and calculates their relevant scores with token-level interaction. For the multimodal question answering,~\model~encodes questions and pages jointly, and produces the textual part and visual part of the multimodal answer through the Decoder and Region Selector. 
\subsection{Input Embeddings}
\model~embeds questions and pages similar to LayoutT5~\cite{visualmrc}. 

\noindent\textbf{Question tokens.} Question $Q$ is tokenized into subword units with SentencePiece~\cite{sentencepiece}. The special token \texttt{</s>} denotes the end of the question.
\begin{align}
    x_Q^\texttt{token}=[q_1, q_2, \dots, q_m, \texttt{</s>}]
\end{align}

\noindent\textbf{Region tokens.} The region tokens consist of a special token $\texttt{<c}_i\texttt{>}$ followed by the OCR words in this region. $\texttt{<c}_i\texttt{>}$ denotes the semantic label of $r_i$.
\begin{align}
    x_{r_i}^\texttt{token}=[\texttt{<c}_i\texttt{>}, w_{i1}, \dots, w_{ik}]
\end{align}

\noindent\textbf{Page tokens.} The sequence of page tokens is the concatenation of all region tokens in the page:
\begin{align}
    x_P^\texttt{token}=[x_{r_1}^\texttt{token};\dots;x_{r_n}^\texttt{token}]
\end{align}

\noindent\textbf{Special embeddings.} Apart from the token embeddings, we add segment embedding $z^\texttt{seg}$ to distinguish question/page tokens.
To incorporate visual and layout information, we add 2D positional embeddings $z^\texttt{pos}$~\cite{layoutlm} and ROI embeddings $z^\texttt{roi}$~\cite{bua} to each page token. 
\begin{align}
    z_Q&=z_Q^\texttt{token}+z^\texttt{seg}_Q \\
    z_P&=z_P^\texttt{token}+z^\texttt{seg}_P+z^\texttt{pos}+z^\texttt{roi}
\end{align}

\subsection{Page Retrieval} \label{sec:page_retrieval}
Page Retrieval aims to find the relevant pages for a question, which requires producing relevant scores between the question and pages. Our~\model~encoder $f$ processes $Q$ and $P$ separately.
\begin{align}
    h_{Q}=f(z_{Q};\theta_f) \\
    h_{P}=f(z_{P};\theta_f)
\end{align}
Since the clues to answer a question usually only appear in a small part of the page, considering the large content of the page, it is difficult for a single global feature to retain detailed clues. Thus, different from general retrieval methods that calculate the cosine similarity between global features, we perform token-level interaction~\cite{filip} between $Q$ and $P$ as shown in Figure~\ref{fig:model}(a). 
Specifically, We calculate the token-level relevant scores $s^{ij}$ between each token in $h^i_{Q}$ and $h^j_{P}$, and aggregate them into two global relevant scores: question-to-page relevant score $S_{Q \rightarrow P}$ and page-to-question relevant score $S_{P \rightarrow Q}$:
\begin{align}
    &s^{ij} = \Vert h^i_{Q} \Vert^\top \Vert h^j_{P} \Vert \\
    &S_{Q\rightarrow P} = \frac{1}{N}\sum_i  \underset{j}{\max}(s^{ij}) \\
    &S_{P\rightarrow Q} = \frac{1}{M}\sum_j \underset{i}{\max}(s^{ij})
\end{align}
We optimize the model by minimizing the NCE loss~\cite{nce} on both the $Q \rightarrow P$ and $P \rightarrow Q$ directions. The loss function for Page Retrieval is written as:
\begin{align}
    &\mathcal{L'}_{(\cdot)} = \frac{1}{B}\sum_i\log\frac{\exp(S_{(\cdot)}^{ii}/\tau)}{\sum_j \exp(S_{(\cdot)}^{ij}/\tau)} \\
    &\mathcal{L}_{\texttt{PR}} = \frac{1}{2}(\mathcal{L'}_{Q \rightarrow P} + \mathcal{L'}_{P \rightarrow Q})
\end{align}
Where $\tau=0.01$ denotes the temperature parameter of NCE, and $B$ denotes the batch size. Note that since we focus on retrieving pages relevant to a given question during inference, we use the score $S_{Q \rightarrow P}$ to rank the candidate pages. 
\subsection{Multimodal Question Answering}
Compared to finding relevant pages for a question, answering a question requires a stronger understanding of both the question and the multimodal contents of the page. Thus, different from Page Retrieval,~\model~encodes question $Q$ and page $P$ jointly to perform early interaction for Multimodal QA. We get the joint hidden state $H$ as follows:
\begin{align}
    H & = f([z_Q, z_P];\theta_f) 
\end{align}
\subsubsection{Textual-part Answer.} 
As shown in Figure~\ref{fig:model}(b), the~\model~decoder receives $H$ and generates the textual-part of the multimodal answer auto-regressively. We train the model in a teacher-forcing manner by minimizing negative log-likelihood loss as below:
\begin{align}
    \mathcal{L}_\texttt{TA}=-\frac{1}{N}\log p(Y|Y_{<},H)
\end{align}
Where $Y=[y_1,\dots,y_N]$ are the ground truth tokens.
\subsubsection{Visual-part Answer.}
The Region Selector $\texttt{RS}$ selects a set of regions to compose the visual-part of the multimodal answer. \texttt{RS} is implemented as a linear projection layer followed by a sigmoid activation. The encoder hidden states corresponding to the $\texttt{<c}_i\texttt{>}$ token is chosen to decide whether $r_i$ is relevant to the question. We minimize the BCE loss to train the model as follows:
\begin{align}
    & p_i = \texttt{RS}(H_{\texttt{<c}_i\texttt{>}};\theta_{\texttt{RS}}) \\
    & \mathcal{L}_\texttt{VA} = -\frac{1}{N}\sum_i y_i \log (p_i)+(1-y_i)\log (1-p_i)
\end{align}
Where $y_i=\{0,1\}$ denotes whether the region $r_i$ belongs to the ground truth vision-part answer.
\subsection{Multitask Learning}
Finally, \model~is optimized in a multitask learning manner, where the final loss function is calculated as follows:
\begin{align}
    \mathcal{L_\texttt{\model}} = \mathcal{L_\texttt{PR}} + \mathcal{L_\texttt{TA}} + \mathcal{L_\texttt{VA}}
\end{align}

\begin{table*}[htpb]
\centering
\begin{tabular}{@{}c|lcccccccccc@{}}
\hline
\multirow{2}{*}{} & \multirow{2}{*}{Model} & \multicolumn{3}{c}{Page Retrieval} & \multicolumn{4}{c}{Textual-part Answer} & \multicolumn{3}{c}{Visual-part Answer} \\ 
 &  & R@1 & R@3 & R@5 & B4 & M & R-L & C & P & R & F1 \\ \hline
1 & PR$_{g}$ & 39.0 & 61.2 & 71.3 & - & - & - & - & - & - & - \\
2 & PR & 80.3 & 93.5 & 95.8 & - & - & - & - & - & - & - \\
3 & PR$_{g}$+TA & 38.3 & 60.8 & 70.4 & 41.5 & 31.8 & 57.4 & 345.3 & - & - & - \\
4 & PR+TA & 80.7 & 93.0 & 95.6 & 42.4 & 32.4 & 58.5 & 355.3 & - & - & - \\
5 & \model~(PR+TA+VA)  & \textbf{81.8} & \textbf{94.4} & \textbf{96.4} & \textbf{42.9} & \textbf{33.0} & 59.5 & 361.6 & \textbf{81.1} & 56.6 & 66.7 \\ 
6 & 3 Single & 80.3 & 93.5 & 95.8 & 42.4 & 32.4 & \textbf{59.6} & \textbf{367.7} & 75.7 & \textbf{60.1} & \textbf{67.0}  \\ \hline
\end{tabular}
\caption{Comparing the URA model with several baselines on Page Retrieval and Multimodal QA.}
\label{tab:expr_sep}
\end{table*}

\begin{table}[htpb]
\centering
\begin{tabular}{@{}p{34pt}m{13pt}m{13pt}m{16.5pt}m{19pt}m{16pt}m{16pt}p{0pt}@{}}
\hline
\multirow{2}{*}{Model} & \multicolumn{4}{c}{Textual-part Answer} & \multicolumn{3}{c}{Visual-part Answer} \\ 
& B4 & M & R-L & C & P & R & F1 \\ \hline
3 Single & 38.0 & 29.6 & 55.1 & 323.9 & 76.9 & \textbf{50.1} & 60.7 \\
\model & \textbf{38.9} & \textbf{30.3} & \textbf{55.5} & \textbf{324.6} & \textbf{82.5} & 48.3 & \textbf{61.0} \\ \hline
\end{tabular}
\caption{Evaluate Multimodal QA under the cascade setting.}
\label{tab:cascade}
\end{table}

\begin{table}[htbp]
\end{table}

\begin{table}[t]
\scalebox{0.95}{
\begin{tabular}{@{}p{57.5pt}m{0pt}m{9pt}m{16.5pt}m{20pt}m{9pt}m{9pt}p{0pt}@{}}
\hline
\multirow{2}{*}{Region} & \multicolumn{4}{p{85pt}}{\hspace{15pt}{Textual-part Ans.} 
 } & \multicolumn{3}{p{65pt}}{Visual-part Ans.} \\ 
 & \multicolumn{1}{c}{B4} & M & R-L & C & P & R & F1 \\ \hline
Text & \multicolumn{1}{c}{43.1} & 33.2 & 59.8 & 364.9 & 84.6 & 72.6 & 78.2 \\
Title & \multicolumn{1}{c}{38.8} & 30.6 & 57.6 & 366.6 & 68.4 & 44.8 & 54.1 \\
Product image & \multicolumn{1}{c}{36.5} & 29.2 & 55.2 & 305.9 & 58.1 & 4.5 & 8.3 \\
Illustration & \multicolumn{1}{c}{37.7} & 30.1 & 58.9 & 332.6 & 32.6 & 3.0 & 5.4 \\
Table & \multicolumn{1}{c}{36.8} & 29.1 & 50.4 & 271.1 & 57.4 & 15.8 & 24.8 \\
Graphic & \multicolumn{1}{c}{32.9} & 27.1 & 51.3 & 271.2 & 60.8 & 16.2 & 25.5 \\ \hline
\end{tabular}
}
\caption{Multimodal QA results on each semantic region.}
\label{tab:mqa_each_category}
\end{table}

\begin{table}[htpb]
\centering
\begin{tabular}{lc}
\hline
Model & MOS \\ \hline
TA+random region & 2.11 \\
TA+nearst region & 2.67 \\
\model & 3.52 \\
Human & 4.70 \\ \hline
\end{tabular}
\caption{Mean Opinion Score of the human evaluation.}
\label{tab:human_eval}
\end{table}

\section{Experiments}
We conduct experiments to validate our~\model~model on the proposed~\dataset~dataset.

\subsection{Evaluation Setup}
\subsubsection{Evaluation settings.} As mentioned before, MPMQA can be naturally split into two subtasks. Thus, we design two evaluation settings for subtask II, Multimodal QA: 1) \textit{separate setting}: evaluating QA given the ground-truth pages; 2) \textit{cascade setting}: evaluating QA given retrieved pages. We adopt the \textit{separate setting} by default if not specified. 
\subsubsection{Evaluation metrics.} For Page Retrieval, we calculate Recall@\{1,3,5\} in each manual, and weigh the scores across manuals by the number of pages. For Textual-part Answer, we report sequence generation metrics BLEU4 (B4)~\cite{bleu}, METEOR (M)~\cite{meteor}, ROUGE-L (R-L)~\cite{rouge} and CIDEr~\cite{cider}. For Visual-part Answer, we report the average Precision (P), Recall (R), and F1 scores on the whole dataset.
\subsection{Baselines} \label{sec:baselines}
We compare our~\model~model with the following baselines:
\begin{itemize}
    \item PR: the Page Retrieval task-specific model. It can conduct the Page Retrieval task only.
    \item PR$_g$: the Page Retrieval task-specific model that uses global features to measure the relevancy between questions and pages. 
    \item PR+TA: the multi-task model that is jointly optimized with Page Retrieval and Textual-part Answer tasks.
    \item PR$_g$+TA: the multi-task model that is jointly optimized with the Page Retrieval (global features) and Textual-part Answer tasks.
    \item 3 Single: 3 separate task-specific models for Page Retrieval, Textual-part Answer, and Visual-part Answer.
\end{itemize}

\subsection{Implementation Details}
We implement the above-mentioned models based on Pytorch~\cite{torch} and Huggingface Transformers~\cite{huggingface}. The encoder and decoder of the models are in standard transformer architecture~\cite{transformer} with T5~\cite{t5} initialization. The models adopt the T5$_{\mathrm{BASE}}$ structure that consists of 12 transformer layers with 768-d hidden states. We train the models for 20 epochs with a batch size of 8 and a learning rate of 3e-5. It takes about 20 hours to converge on 1 NVIDIA RTX A6000 GPU. We choose the model that performs best on the validation set, and report its performance on the test set. We consider the most relevant page for the Multimodal QA task. 
\subsection{Results and Analysis}

\noindent\textbf{Comparing~\model~with several baselines.} Table~\ref{tab:expr_sep} shows the comparison between~\model~and the baselines described above. Comparing row 1 and 2, we observe that retrieving with global features performs much worse than with the token-level interaction method described in the previous section, which indicates that the Page Retrieval task requires fine-grained interaction between questions and pages, since question-related clues usually occur in local area of the page. Additionally, jointly optimizing TA with PR$_g$ (row 3) hurts both TA (row 6) and PR$_g$ (row 1). In contrast, jointly training TA and PR (row 4) affects each other less. It may be because that question answering task requires fine-grained understanding, and it does not conflict with the token-level interaction in PR. Finally, jointly training with VA also helps both the Page Retrieval and TA task (row 5). Compared to the 3 task-specific models,~\model~achieves even better performance over Page Retrieval and Multimodal QA.

\noindent\textbf{Multimodal QA under cascade setting.} Table~\ref{tab:cascade} shows that~\model~also outperforms multiple task-specific models under the cascade setting. However, we observe a large performance gap between the separate setting and the cascade setting. Considering that the cascade setting is closer to real applications, the Page Retrieval task could be the bottleneck of the MPMQA. Thus, investigating more powerful retrieval models, or models that can directly answer questions from the whole manual will benefit the MPMQA task.

\noindent\textbf{Multimodal QA broken down by Semantic Regions.}
Table~\ref{tab:mqa_each_category} shows the~\model~performance breaking down in semantic categories.~\model~performs well on text regions, but worse on visually-rich regions such as product images, illustrations, tables, and graphics, which indicates that a more powerful multimodal understanding ability is required to achieve better performance in the MPMQA task.

\noindent\textbf{Human evaluation.} We conduct a human evaluation to verify whether the multimodal answer is helpful for user understanding. We sample 50 question-answer pairs from the test set of PM209. We inference the TA task-specific model and~\model~on this subset to get the Text-only Answer and Multimodal Answer respectively. Considering the annotators may easily distinguish the two models according to whether there are visual-part outputs, we attach visual-part outputs to the text-only answers with two baseline approaches: 1) random region: randomly selecting two regions from the page; 2) nearest region: selecting the neighbor region that shares the most OCR words with the textual answer. We provide the question and four answers simultaneously to 20 human evaluators, and ask them to rate each answer by 1-5 points according to whether the answer is helpful to address the given question. The four answers include: two text-only answers attached with visual-part outputs, the multimodal answer produced by our model, and the ground truth multimodal answer by humans. 
The Mean Opinion Score (MOS) of the 4 answers is shown in Table~\ref{tab:human_eval}. It shows that the multimodal answer produced by~\model~are more helpful than text-only answers.

\section{Conclusion}
In this paper, we propose the Multimodal Product Manual Question Answering (MPMQA) task, which requires the model to comprehend multimodal content in an entire product manual and answer questions with multimodal outputs. To support the MPMQA task, we construct the large-scale dataset~\dataset~with human annotations. It contains 22,021 multimodal question-answering pairs on 209 product manuals across 27 well-known consumer brands. The multimodal answer to each question consists of a textual-part in natural language sentences, and a visual-part consisting of regions from the manual. For the MPMQA task, we further propose a unified model that retrieves relevant pages and generates multimodal answers based on multitask learning. It achieves competitive results compared to multiple task-specific models. We release the dataset, code, and model at https://github.com/AIM3-RUC/MPMQA. 

\section*{Acknowledgements}
This work was partially supported by the National Natural Science Foundation of China (No. 62072462) and the National Key R\&D Program of China (No. 2020AAA0108600).
\balance

\bibliography{aaai23}
\clearpage
\appendix
\onecolumn
\section{Product Manual Collection}
\subsection{Data Sources}
The product manuals in~\dataset~are from two sources: 1) \textbf{E-manual corpus}~\cite{emanual}; 2) \textbf{official websites} of the brands. 

\subsubsection{Source 1: E-manual corpus}
E-manual corpus~\cite{emanual} is a large-scale text corpus. It is constructed by crawling product manuals from the website\footnote{www.manualsonline.com}
and extracting their text contents. 
Its metadata includes the product categories and brands of the manuals. We find that the original files of the E-manual corpus are in PDFs and a part of them are suitable for the MPMQA task. Thus, we download the original PDF files and select 139 manuals to be included in ~\dataset~based on the following rules:
\begin{enumerate}
    \item Manuals must be in well-known electronic brands and categories. 
    \item Manuals must be born-in-digital.
    \item Manuals must be more than 10 pages in length. 
    \item For the product in the same series, keep the longest one.
    \item Manuals must contain visually-rich contents. This filtering process is done manually.
\end{enumerate}
The 139 manuals from the E-manual corpus cover 9 brands and 73 categories as shown in Figure~\ref{fig:brands_and_products_emanual}.

\subsubsection{Source 2: Official websites}
\begin{figure}[h]
    \centering
    \begin{subfigure}{0.495\linewidth}
    \centering
    \includegraphics[width=1\linewidth]{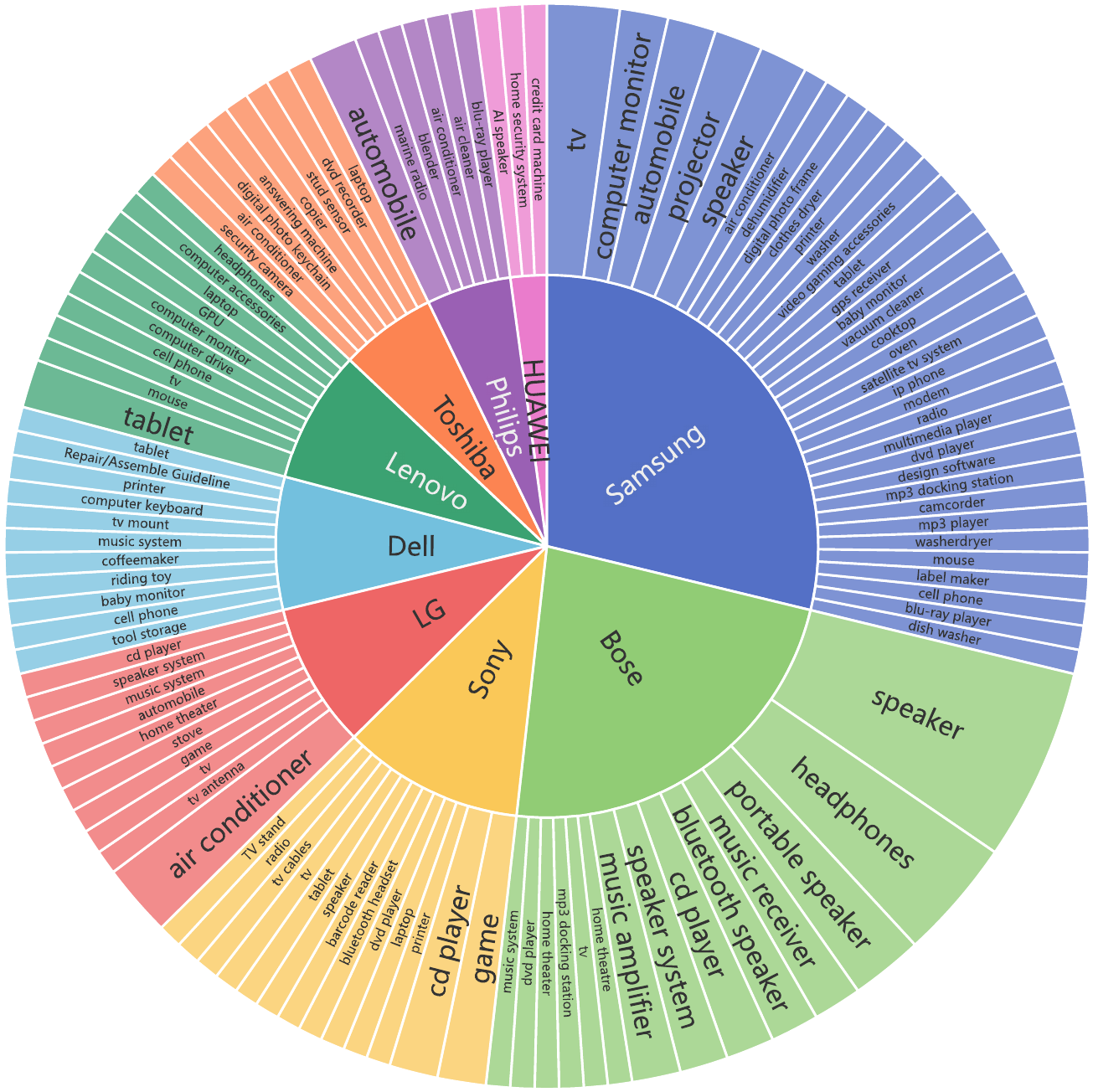}
    \caption{E-manual (139 manuals).}
    \label{fig:brands_and_products_emanual}
    \end{subfigure}
    \hfill
    \begin{subfigure}{0.495\linewidth}
    \centering
    \includegraphics[width=1\linewidth]{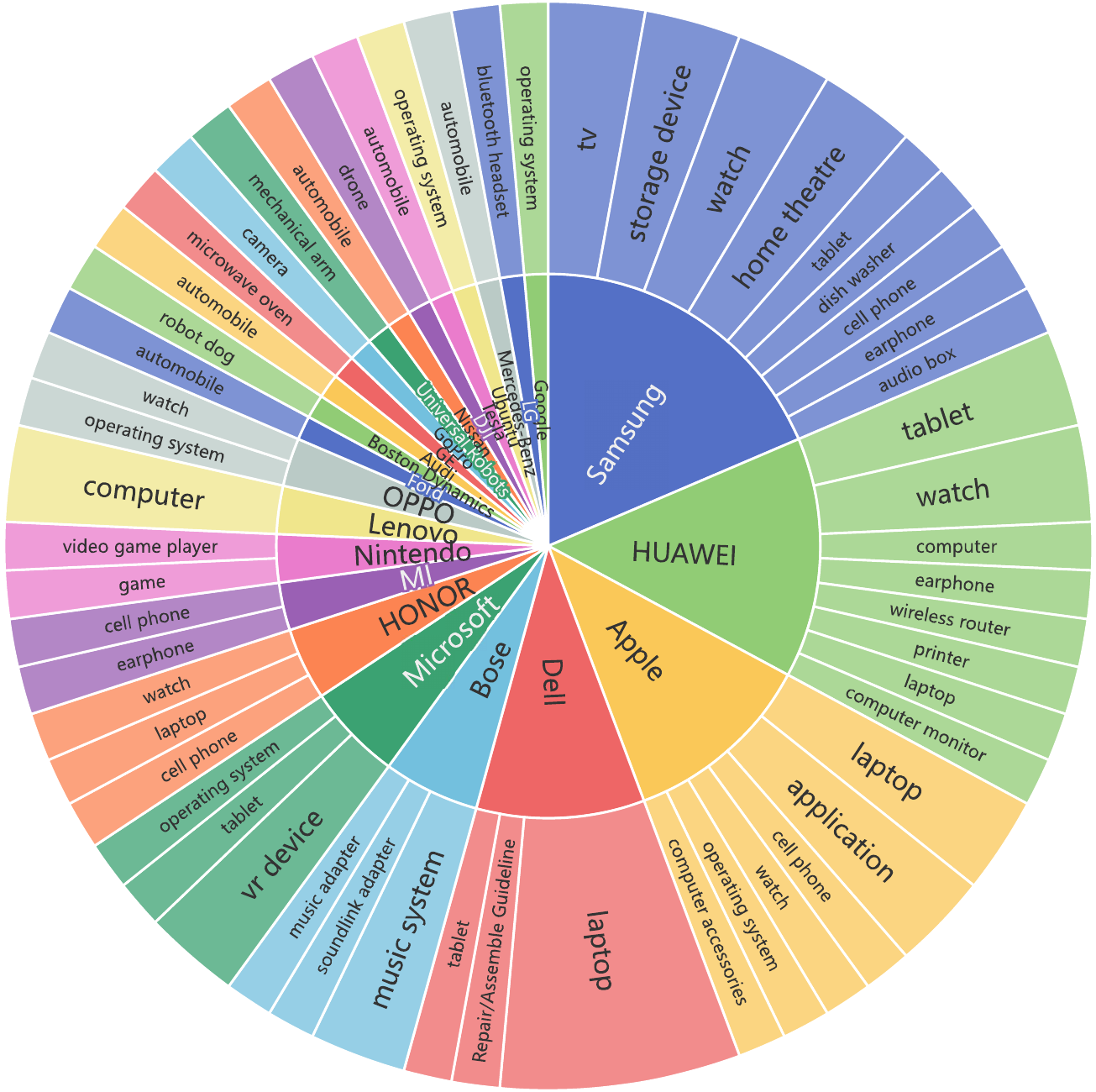}
    \caption{Official websites (70 manuals).}
    \label{fig:brands_and_products_official}
    \end{subfigure}
    \caption{Brands and products of the manuals from two sources.}
\end{figure}

Tough the E-manual corpus contains a large number of product manuals, we find that it covers limited product categories. For example, it does not contain recently popular products such as cell phones, smart watches, drones, software (operating systems, games, applications) and hardware (storage devices, robots), etc. To make the~\dataset~closer to the real-world application, we further collect 70 manuals from the official websites of major brands including Samsung, HUAWEI, Apple, etc. The data filtering process is consistent with that in E-manual. These 70 manuals cover 24 brands and 31 categories, as shown in Figure~\ref{fig:brands_and_products_official}.

\subsection{LDA analysis}
To further verify the diversity of the~\dataset~dataset, we analyze the topics of the product manuals in~\dataset~with Latent Dirichlet Allocation (LDA)~\cite{lda}. We extract 10 topics from all manuals. Table~\ref{tab:lda_topics} shows that~\dataset~covers diverse topics including PC, automobiles, speakers, applications, hardware, cellphones, cameras, etc. The topics of~\dataset~are much richer than S10 QA and Smart TV/Remote QA~\cite{emanual} which only focus on one specific product. Topics in~\dataset~are also very different from VisualMRC~\cite{visualmrc} which consists of open-domain webpages and contains rare topics about specific consumer products.
\begin{table}[h!]
    \centering
    \scalebox{1.0}{
    \begin{tabular}{c|l}
        \toprule
        No. & Topic words\\ \midrule
        1 & app windows tap  microsoft account network update users \\
        2 & vehicle seat assist driving brake charge tire child \\
        3 & speaker remote bose supported samsung bluetooth video network \\
        4 & click text table add format cell image app \\
        5 & huawei app tap charging speed tire wifi enable \\
        6 & signal channel field parameters surround delay filter cycle \\
        7 & assembly board key working drive base keyborad step \\
        8 & tap iphone app watch swipe options network contacts \\
        9 & disc windows usb video player file lenovo memory \\
        10 & air remote speed camera indoor filter controller appliance \\ \bottomrule
    \end{tabular}
    }
    \caption{Topic words extracted through LDA.}
    \label{tab:lda_topics}
\end{table}

\section{Examples of Semantic Regions}
\begin{figure*}[ht]
    \centering
    \includegraphics[width=\linewidth]{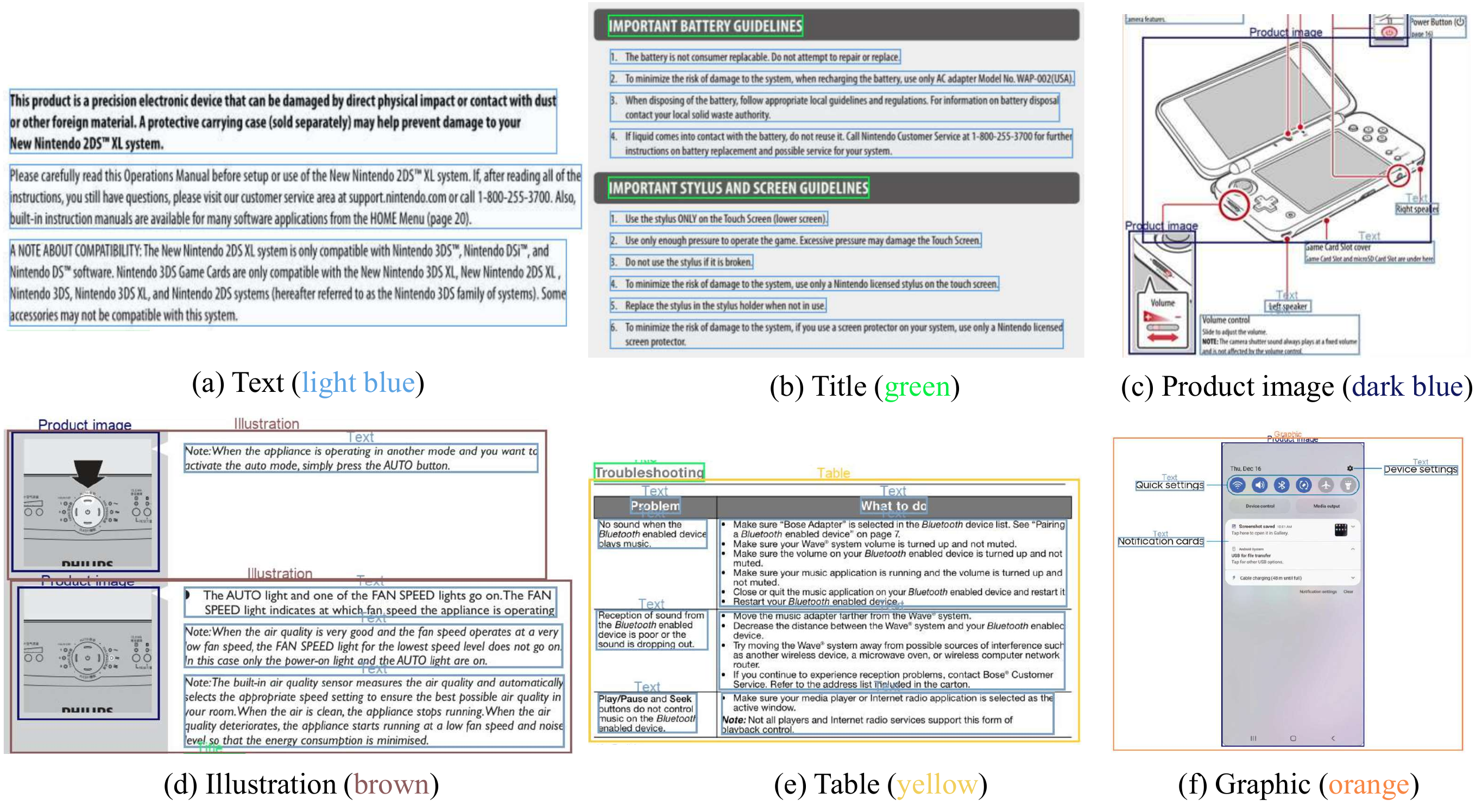}
    \caption{Examples of the six semantic regions. Better to read in color.}
    \label{fig:region_cases}
\end{figure*}

\begin{figure*}[ht]
    \centering
    \includegraphics[width=\linewidth]{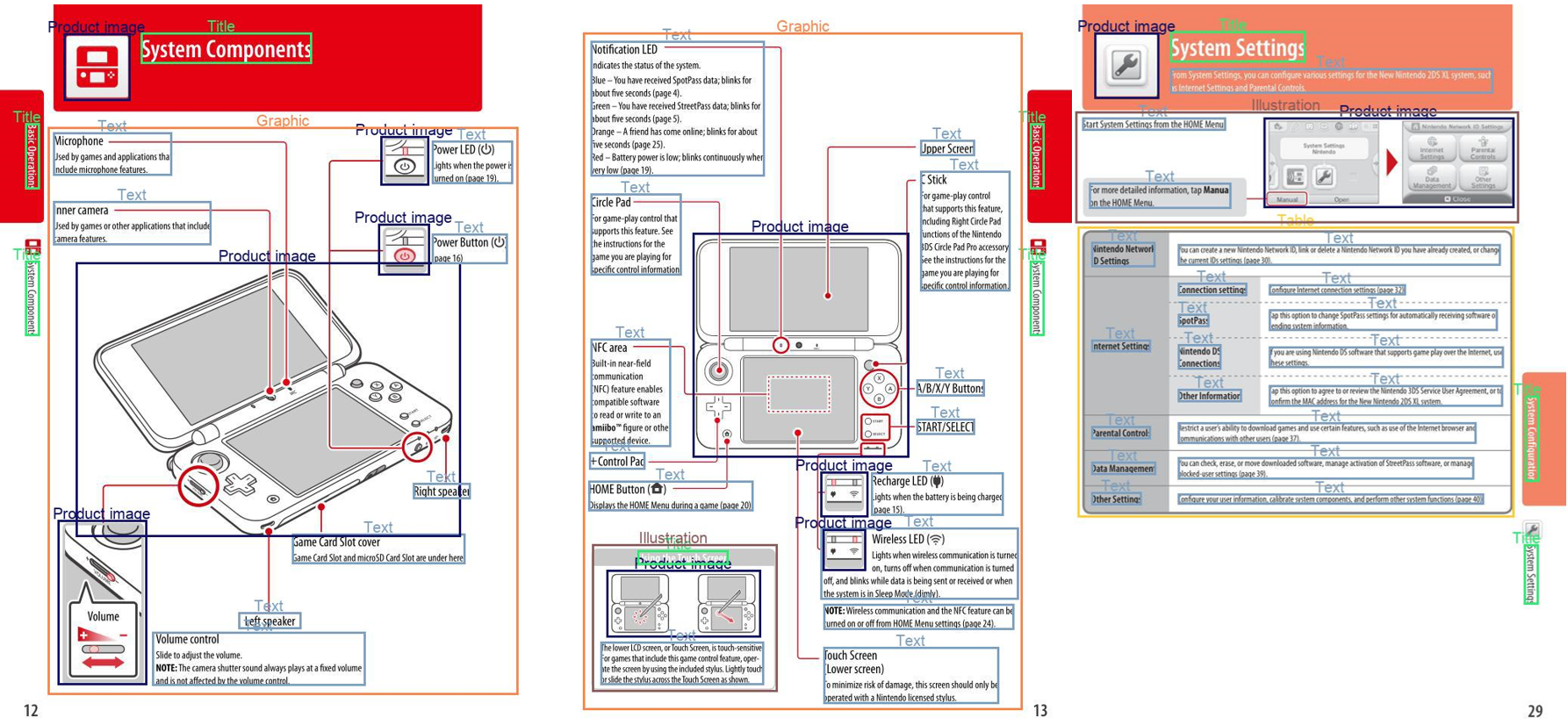}
    \caption{Examples of semantic regions in real pages. Better to read in color.}
    \label{fig:real_page}
\end{figure*}

Figure~\ref{fig:region_cases} shows some examples of each semantic region, and Figure~\ref{fig:real_page} shows an example of the semantic regions in real pages. Note that region overlaps are allowed considering the complexity of the layout in product manuals. 

\section{Answers that across multiple pages}
During the QA annotation, we allow the crowd workers to pose questions that require content across multiple pages to answer. However, as shown in Table~\ref{tab:qa_relevant_pages}, we find that only 1.59\% questions are relevant with more than one page. We omit these questions during model training, but keep their annotation for future research. 

\begin{table}[ht]
\centering
\begin{tabular}{@{}l|ccccc|c@{}}
\toprule
Relevant pages & 1 & 2 & 3 & 4 & 5 & Total \\ \midrule
QA pairs & 21670 & 330 & 18 & 1 & 2 & 22021 \\ \bottomrule
\end{tabular}
\caption{The number of QA pairs with a specific number of relevant pages.}
\label{tab:qa_relevant_pages}
\end{table}

\section{Interface of the Human Evaluation}
\begin{figure}[!h]
    \centering
    \includegraphics[width=1\linewidth]{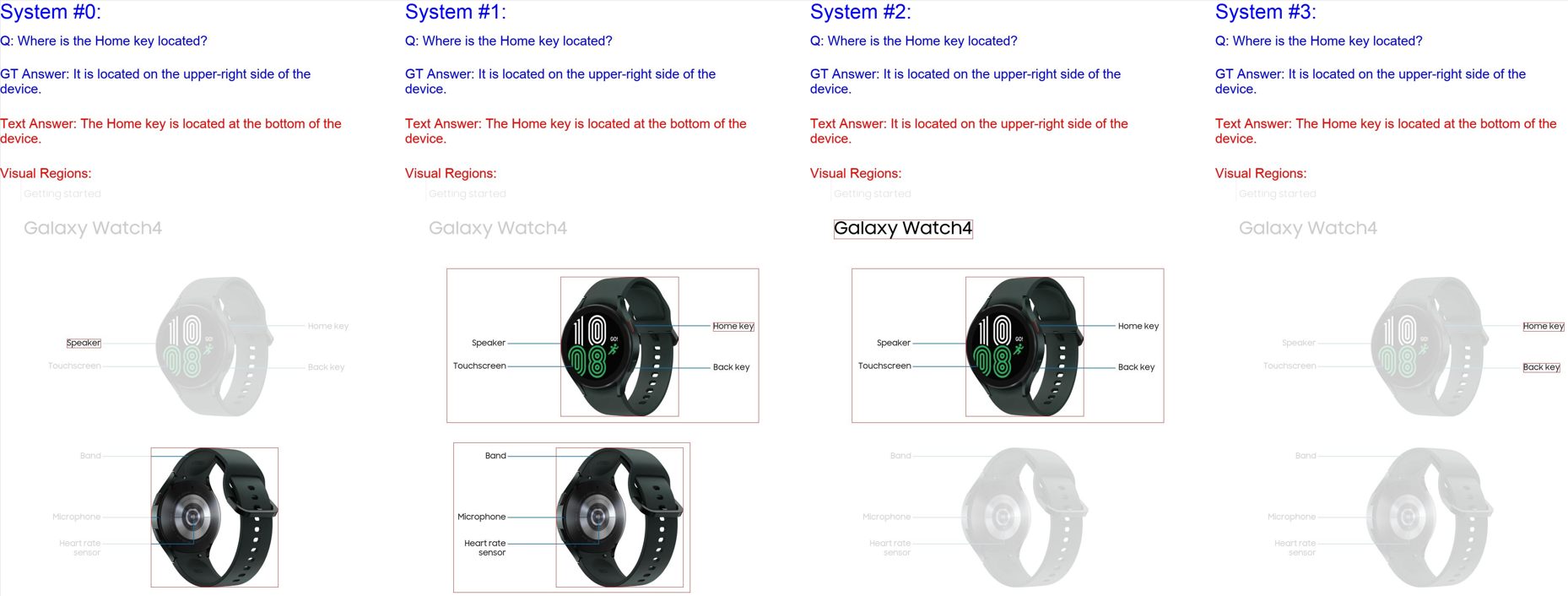}
    \caption{A screenshot of the user interface in Human Evaluation.}
    \label{fig:human_eval_interface}
\end{figure}

Figure~\ref{fig:human_eval_interface} shows the interface of Human Evaluation. The human evaluators are provided with the four randomly shuffled answers simultaneously. They are asked to score each answer. We present the visual-part answers of each system by highlighting the regions on the pages. 

\section{Case Study}
\begin{figure}[htpb]
    \centering
    \includegraphics[width=\linewidth]{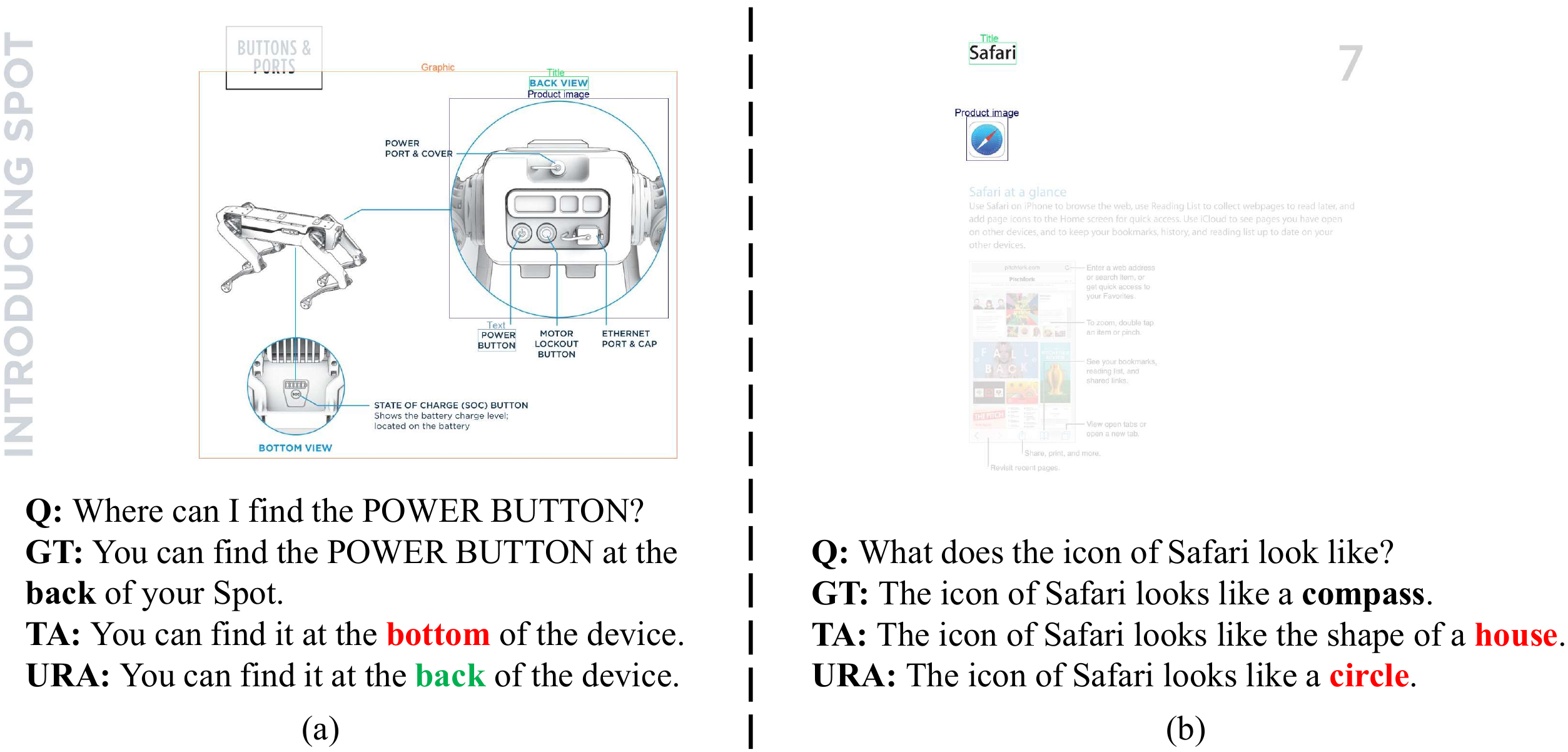}
    \caption{Cases of multimodal question answering. GT refers to the ground truth textual-part answer.}
    \label{fig:qa_cases}
\end{figure}

Figure~\ref{fig:qa_cases} shows two inference cases of the~\model~model compared with the TA baseline and ground truth answers. The visual-part answers of~\model~are highlighted in the images. In case (a), given the question 'Where can I find the POWER BUTTON',~\model~is able to generate the correct answer. In contrast, the tasks-specific model TA fails. This is because the region 'BACK VIEW' is the only information source of the answer, and the joint learning with visual-part answers in~\model~helps it to attend to this region. In case (b), given the question of 'What does the icon of Safari look like?', both~\model~and TA does not produce a satisfactory textual answer. However, the URA model could predict the correct regions as the visual-part answer. It thus can directly show these regions to users to answer this question. 

\end{document}